\documentclass{article}
\usepackage{ijcai25}

\usepackage{times}
\usepackage{soul}
\usepackage{url}
\usepackage[hidelinks]{hyperref}
\usepackage[utf8]{inputenc}
\usepackage[small]{caption}
\usepackage{graphicx}
\usepackage{amsmath}
\usepackage{amsthm}
\usepackage{booktabs}
\usepackage{algorithm}
\usepackage{algorithmic}
\usepackage[switch]{lineno}

\usepackage{hyperref}
\hypersetup{
    colorlinks=true,
    linkcolor=blue,
    urlcolor=cyan,
    pdfpagemode=FullScreen,
}
\usepackage{colortbl} 
\usepackage{multirow} 
\usepackage{booktabs} 
\usepackage{amsmath}    
\usepackage{amssymb}    
\usepackage{amsfonts}   
\usepackage{comment}
\usepackage{xcolor}
\definecolor{customgreen}{HTML}{00C474} 
\definecolor{customred}{HTML}{ff6768}   
\definecolor{ourworkclr}{HTML}{6495ED}

\definecolor{bestresult}{HTML}{E69191} 
\definecolor{secondbestresult}{HTML}{EDB8B0}   

\newcommand{\capourworkclr}[1]{\textcolor{ourworkclr!30}{\rule{#1}{10pt}}}   
\newcommand{\capbestresult}[1]{\textcolor{bestresult}{\rule{#1}{10pt}}}      
\newcommand{\capsecondbestresult}[1]{\textcolor{secondbestresult}{\rule{#1}{10pt}}}  
\newcommand{\supp}[1]{%
  \textcolor[HTML]{FFD700}{#1}%
}

\title{Exemplar-condensed Federated Class-incremental Learning}

\author{
Rui Sun, Yumin Zhang, Varun Ojha, Tejal Shah, Haoran Duan, Bo Wei, Rajiv Ranjan
\affiliations
School of Computing, Newcastle University, UK
\emails
\{Rui.Sun, Y.Zhang361, Varun.Ojha, Tejal.Shah, Haoran.Duan, Bo.Wei, Rajiv.Ranjan\}@newcastle.ac.uk
}

\begin{document}

\maketitle

\begin{abstract}
    We propose \textbf{E}xemplar-\textbf{Co}ndensed fede\textbf{r}ated class-increment\textbf{a}l \textbf{l}earning (ECoral) to distill the training characteristics of real images from streaming data into informative rehearsal exemplars. The proposed method eliminates the limitations of exemplar selection in replay-based approaches for mitigating catastrophic forgetting in federated continual learning (FCL). The limitations are particularly related to the heterogeneity of information density of each summarized data. Our approach maintains the consistency of training gradients and the relationship to past tasks for the summarized exemplars to represent the streaming data compared to the original images effectively. Additionally, our approach reduces the information-level heterogeneity of the summarized data by inter-client sharing of the disentanglement generative model. Extensive experiments show that our ECoral outperforms several state-of-the-art methods and can be seamlessly integrated with many existing approaches to enhance performance.
\end{abstract}

\section{Introduction}

Federated Learning (FL)~\cite{mcmahan2017communication} enables decentralized training across clients, addressing data silos and privacy concerns, with applications in areas like smart healthcare\cite{nguyen2022federated} and IoT~\cite{nguyen2021federated,jiang2024blockchained}. However, traditional FL assumes static data, which conflicts with scenarios where new classes emerge~\cite{yoon2021federated,ma2022continual}. Finetuning a pre-trained model on novel data leads to \textit{catastrophic forgetting}~\cite{li2017learning}, and limited storage and privacy restrictions make retraining impractical.

\begin{figure}[!t]
  \centering
  \includegraphics[width=\linewidth]{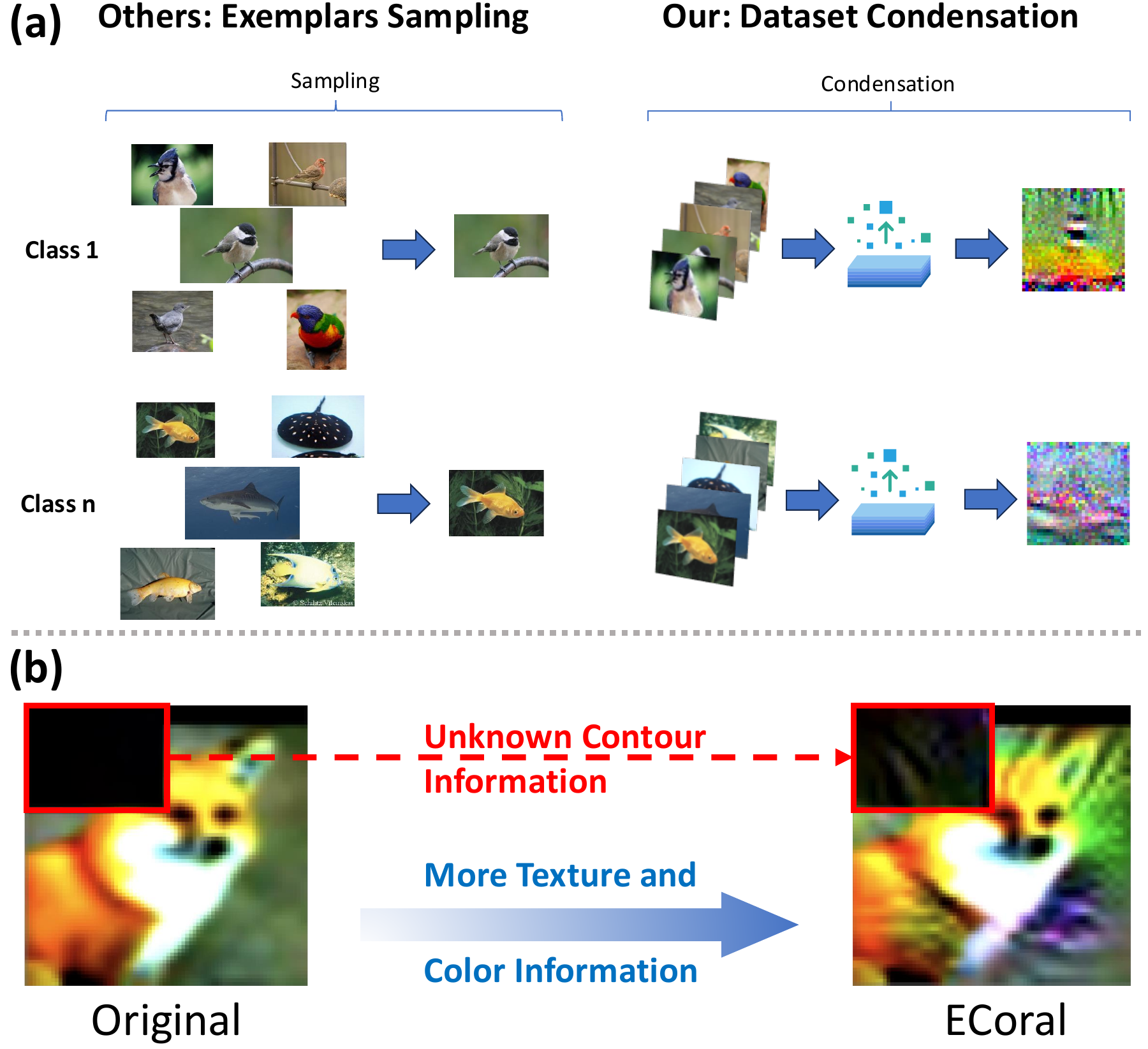}
  \caption{Comparison with other approaches: (a) Most FCL methods rely on exemplars sampled from training data, while ECoral extracts more comprehensive, informative exemplars. (b) ECoral captures hidden contours and enriches class-specific features like texture and color, making exemplars more representative of the data.}
  \label{fig:intro_setting}
\end{figure}

Recent work~\cite{dong2022federated,zhang2023target} has enabled incremental learning of new classes in FL, known as Federated Class-Incremental Learning (FCIL). Among these, rehearsal-based methods~\cite{dong2022federated,qi2023better} store and replay exemplars from prior tasks to reduce forgetting. However, storage constraints and privacy concerns limit data storage, raising the question: \textit{How can a restricted dataset be used to capture more information and prevent forgetting without compromising privacy?}

Data condensation (DC)~\cite{wang2018dataset,zhao2020dataset} techniques like distribution/feature matching\cite{wang2022cafe,zhao2023dataset} or gradient matching~\cite{zhao2020dataset,cazenavette2022dataset} have emerged to synthesize compact datasets. These methods preserve essential characteristics of the original data, enabling models trained on condensed datasets to perform similarly to those trained on full datasets. Several works~\cite{goetz2020federated,hu2022fedsynth,xiong2023feddm} have applied DC in FL, replacing model parameter exchanges with synthetic data. However, DC in FL faces challenges like \textit{meta-information heterogeneity}, where summarizing non-IID data from clients can disrupt optimization and degrade performance.

To address this, we propose the Exemplar-\textbf{Co}ndensed fede\textbf{r}ated class-increment\textbf{a}l \textbf{l}earning (ECoral) framework. ECoral uses a dual-distillation approach: one for extracting informative exemplars (e.g., contours, textures, and color), as illustrated in Figure~\ref{fig:intro_setting}, and one for preserving prior knowledge to avoid forgetting. Our method enhances privacy by shifting from raw data storage to condensed exemplars. Experimental results show ECoral outperforms baselines, achieving 49.17\% accuracy on CIFAR-100 with 10 tasks, and 32.42\% with 50 tasks, surpassing the best baseline by 5.32\% and 10.29\%, respectively.

\section{Related Work}
\subsection{Dataset Condensation}
Dataset Condensation (DC) reduces dataset size while maintaining enough information for models to perform similarly to training on the full dataset. Early works \cite{wang2018dataset,zhao2020dataset} framed this as a bi-level optimization problem, aiming to create a smaller dataset that preserves key characteristics of the original data. Techniques focus on matching distributions, features, or gradients between the original and condensed data. DC has been used in continual learning to mitigate catastrophic forgetting by compressing past task data into small, representative memory sets \cite{masarczyk2020reducing,gu2024summarizing}.

In FL, DC helps reduce communication overhead by transmitting condensed datasets instead of full models or gradients \cite{goetz2020federated,hu2022fedsynth}. For example, \cite{liu2022meta} applied DC to handle data heterogeneity in FL, reducing communication costs and bias. These methods show DC’s potential in addressing FL’s resource and privacy constraints, allowing synthetic data sharing while preserving client privacy. Our approach extends DC to federated class-incremental learning (FCIL), improving exemplar informativeness and mitigating catastrophic forgetting in non-IID scenarios.

\subsection{Federated Continual Learning}
Federated Continual Learning (FCL) extends traditional FL to dynamic environments, where new tasks or classes emerge incrementally. Unlike static FL, FCL must handle evolving data, catastrophic forgetting, non-IID distributions, and communication constraints \cite{yoon2021federated}.

Rehearsal-based methods are common for addressing catastrophic forgetting in FCL. These methods store a limited number of exemplars from previous tasks and replay them during training on new tasks, helping the model retain knowledge of older classes \cite{dong2022federated,qi2023better}. For example, GLFC \cite{dong2022federated} uses sample reconstruction, while FedCIL \cite{qi2023better} employs generative models for replay. However, these methods face challenges in federated settings due to privacy and storage limitations. TARGET \cite{zhang2023target}, an exemplar-free distillation method, offers a privacy-preserving alternative by using knowledge distillation from global models and generating synthetic data, reducing reliance on real data storage while addressing forgetting in non-IID settings.

A key limitation of rehearsal-based methods is class imbalance in exemplar memory. Since clients usually have data for only some classes, the stored memory is biased towards local classes, exacerbating forgetting and overfitting. This issue is worse in non-IID settings, where the data distribution across clients is highly skewed. Even TARGET struggles with capturing class diversity due to reliance on global distillation.

\section{Preliminaries}
\label{sec:pre}

\noindent{\textbf{Federated Class-incremental Learning.}}
Federated Class-incremental Learning (FCIL) collaboratively trains a global model using streaming data introducing new classes sequentially. Training spans $T$ tasks $\{\mathcal{T}^t\}_{t=1}^T$ with $C$ local clients and a central server $S_g$. Each task has $R$ global communication rounds where a subset of clients trains using the latest global model $\theta^{r,t}$. Each client maintains a fixed-size memory $\mathcal{M}_l$ for knowledge replay, divided into original data ($\mathcal{M}_{\text{orig}}$), condensed exemplars ($\mathcal{M}_{\text{cond}}$), and summary data ($\mathcal{M}_{\text{sum}}$). Clients train using both current task data and replayed memory samples, optimizing:

\begin{equation}
\theta^{r,t}_l = \arg\min_{\theta^{r,t}} \mathcal{L}(\theta^{r,t}; \mathbf{B}_n) + \lambda \mathcal{L}_m(\theta^{r,t}; \mathbf{B}_m),
\end{equation}

where $\mathcal{L}$ and $\mathcal{L}_m$ are the loss functions for current task and memory data, respectively. Locally updated models $\theta^{r,t}_l$ are aggregated at the server to update the global model $\theta^{r+1,t}$. Data distributions across clients are non-IID, with category sets evolving per task, leading to performance challenges in maintaining past knowledge.

\noindent{\textbf{Client increment strategy.}}
To simulate real-world scenarios, we use the client increment strategy from GLFC \cite{dong2022federated}, dividing participants into three groups per task: Old ($\mathcal{G}_o$), In-between ($\mathcal{G}_b$), and New ($\mathcal{G}_n$). Old clients handle only past task data, In-between clients manage both past and current task data, and New clients focus on current task data. Group compositions are dynamically updated, gradually increasing total participants and mimicking streaming data in federated learning.

\subsection{Problem Definition}

\subsubsection*{Forgetting in FCIL}
The global model aims to minimize classification error on current categories $\mathcal{K}_t$, but privacy constraints and resource limits restrict access to past data, exacerbating category imbalance. This leads to catastrophic forgetting, where performance on earlier tasks degrades. The goal is to minimize errors on $\mathcal{K}_t$ while preserving prior knowledge:

\begin{equation}
\min _{\theta_t} \sum_{k \in \mathcal{K}_t} \sum_{i=1}^{N_k} \mathcal{L}\left(\mathbf{P}_l^t\left(\mathbf{x}_{l,i}^t ; \theta_{r,t}\right), \mathbf{y}_{l,i}^t\right).
\end{equation}

\subsubsection*{Meta-information Heterogeneity} \label{sec:meta-info-hetero}
Non-IID data condensation retains non-IID characteristics, causing meta-information heterogeneity. Each client’s condensed dataset $\mathcal{M}_{\text{cond}}$ reflects distinct distributions $\mathbf{P}_l^{\text{cond}}$, often conflicting during global aggregation. This results in suboptimal global performance, quantified by increased global loss $\Delta \mathcal{L} = \mathcal{L}_{\text{non-iid}}(\theta^{r,t}) - \mathcal{L}_{\text{iid}}(\theta^{r,t}) > 0$. Addressing this heterogeneity is crucial for mitigating its negative impact on global model performance.

\section{Exemplar-condensed FCIL with Dual-distillation Structure}

\begin{figure*}[!ht]
  \centering
  \includegraphics[width=\linewidth]{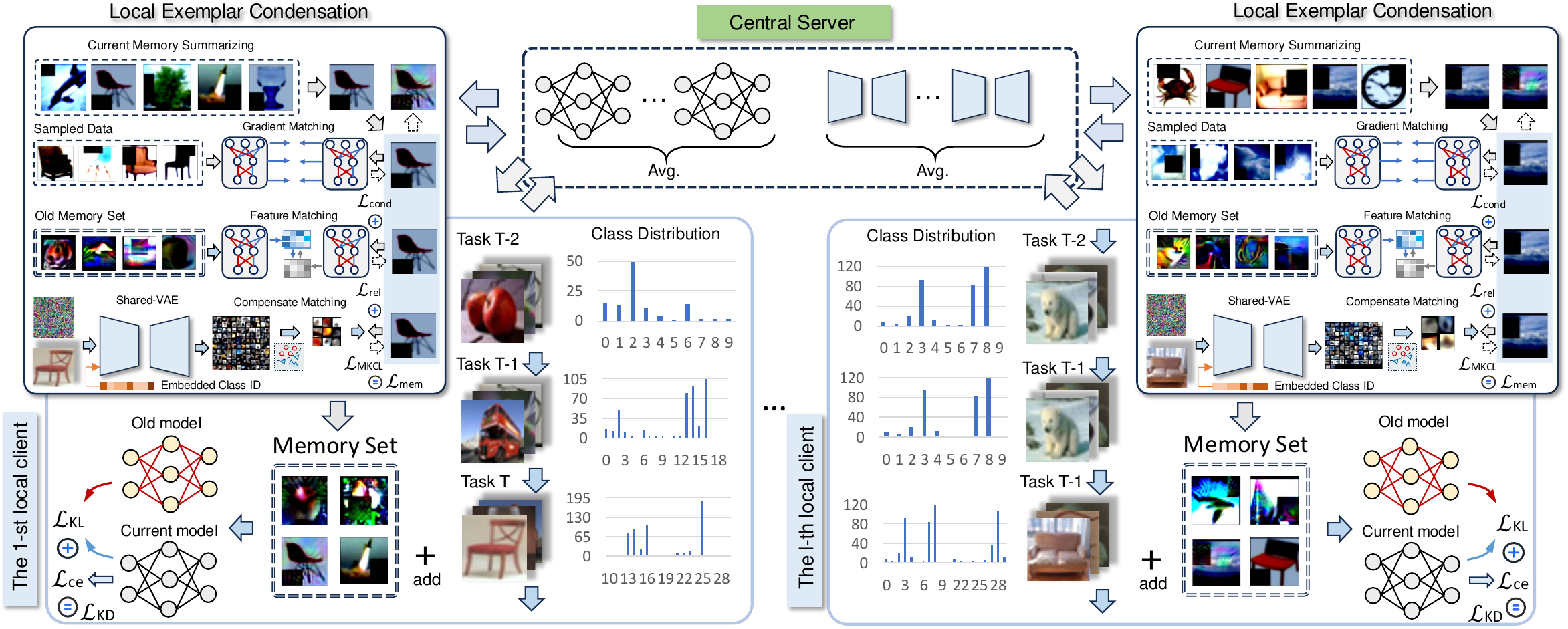}
  \caption{ECoral Overview: Clients continuously learn from new class data sequences using a dual-distillation structure to mitigate catastrophic forgetting. The exemplar condensation process involves three key components: a gradient matching loss (\(\mathcal{L}_{\text{cond}}\)) for meta-information distillation, a feature matching loss (\(\mathcal{L}_{\text{rel}}\)) for consistency between condensed samples and real images, and a compensation loss (\(\mathcal{L}_{\text{MKCL}}\)) to address meta-information heterogeneity using disentangled features from a shared global model (Shared-VAE). A knowledge distillation loss (\(\mathcal{L}_{\text{KD}}\)) helps retain prior knowledge.}
  \label{fig:overall_stru}
\end{figure*}

\subsection{Online Exemplars Condensation}
In edge devices within FCIL, where memory space is highly restricted, most existing approaches focus on efficient exemplar sampling strategies. Compared to these, our approach enhances the meta-knowledge capacity of each individual image, thereby increasing its information level, and also balances these improvements with memory efficiency. The balance and trade-offs are further explained in the following sections, highlighting comparisons and improvements over existing methods.

\subsubsection*{Adjustable Memory.} Efficient management of fixed memory for rehearsal requires sophisticated selection algorithms, as seen in methods like~\cite{rebuffi2017icarl,aljundi2019gradient}. However, these methods are not suitable for distilling meta-knowledge into exemplars from the entire local training dataset.

Unlike conventional online methods like SSD~\cite{gu2024summarizing}, which assume balanced class data and prior knowledge of total classes, our approach overcomes key limitations. SSD allocates only a small fraction of memory to old exemplars, wasting space on current data. In FL, this issue is amplified by non-IID data and class imbalance, as clients typically hold subsets of the full class set.

To address this, we propose a dynamic memory allocation strategy. At the start of each task, we adjust number of stored exemplars per class. For example, if the first task has 10 classes and the memory space is 100 exemplars, we store 10 per class. When a new task introduces 10 more classes, we reduce the previous exemplars to 5 and allocate 50 slots to new classes, ensuring balanced memory and efficient rehearsal.

\subsubsection*{Meta-knowledge Condensation via Gradient Matching.} 
In Federated Class Incremental Learning (FCIL), where memory is limited, our approach goes beyond exemplar selection to enhance each image's information capacity, optimizing memory for effective rehearsal. The goal of Meta-Knowledge Condensation is to minimize the divergence between the memory set and the local task's data distribution, resulting in an optimized memory set, $\mathcal{\hat{M}}_l$. We hypothesize that a global model trained on condensed exemplars can perform similarly to one trained on the full local dataset.

For the $t$-th task, let $\mathcal{M}_k$ represent condensed exemplars for class $k$, with size $m$. The mini-batch data for class $k$ is $\mathbf{B}_k$. The objective is to reduce the divergence between $\mathcal{M}_k$ and $\mathbf{B}_k$. Drawing on dataset condensation methods~\cite{zhao2020dataset}, we use a gradient-based metric to align the model updates on condensed samples and original data, refining exemplars to approximate the full dataset's performance. The gradient matching objective is:

\begin{equation}
\mathcal{L}_{\text{cond}} = f_{\text{dist}} \left( \nabla_{\theta^{r,t}} \mathcal{L}_{ce}(\theta^{r,t}; \mathcal{M}_k), \nabla_{\theta^{r,t}} \mathcal{L}_{ce}(\theta^{r,t}; \mathbf{B}_k) \right)
\end{equation}
where $f_{\text{dist}}$ is a distance function.

\subsection{Current Knowledge Condensation via Feature Matching.} 
In conventional dataset condensation~\cite{zhao2020dataset}, gradient matching alternates with model updates to simulate comprehensive training and match gradients across stages. When a new batch $\mathbf{B}_n$ arrives, the condensation model’s parameters $\omega$ are updated as:

\begin{equation}
\omega \leftarrow \omega - \eta \nabla_{\omega}\mathcal{L}_{ce}(\omega; \mathbf{B}_n),
\end{equation}

where $\eta$ is the learning rate. Only real images are used in this update to prevent knowledge leakage.

In continual learning, the number of classes changes, so we adapt the SSD method~\cite{gu2024summarizing} by re-initializing the model when new classes appear. To preserve gradient information for both current and prior classes, we update the model using both current data and stored images $\mathcal{M}_{\text{orig}}$:

\begin{equation}
\omega \leftarrow \omega - \eta \nabla_{\omega}\left( \mathcal{L}_{ce}(\omega; \mathbf{B}_n) + \mathcal{L}_{ce}(\omega; \mathcal{M}_{\text{orig}})\right)
\end{equation}

Next, we use a relationship-matching strategy to ensure consistency between condensed samples and real images. The matching objective is:

\begin{multline}
\mathcal{L}_{\text{rel}} = f_{\text{dist}}\left(\rho(\mathcal{M}_k, \mathcal{M}_{\text{cond}} \setminus \mathcal{M}_k, \omega), \right. \\
\left. \rho(B_k, \mathcal{M}_{\text{cond}} \setminus \mathcal{M}_k, \omega)\right),
\end{multline}

where $\rho$ computes feature relationships, and $\Phi$ is the feature extraction function.

To address meta-information heterogeneity, we tackle data quantity and class shifts in a privacy-preserving way.

\subsubsection*{Client-wise Feature Disentanglement.}
To handle feature and class skew, we enable each client to generate features from both its own and other clients' data, addressing local class biases. Using a Shared-VAE model~\cite{burgess2018understanding,higgins2017beta}, we generate features from random noise for unseen classes. At the start of each local update, the encoder and decoder are updated with global parameters, and disentangled features are generated for local or unseen classes, represented as $\mathbf{H}$.

\subsubsection*{Unbiased Representative Feature Prototypes.}
A globally trained Shared-VAE tends to favor the majority class distribution. To counteract this, we use unbiased feature prototypes for each class. We apply the FINCH clustering algorithm~\cite{sarfraz2019efficient} to cluster class-specific features, generating representative prototypes for each class. Each class’s feature set is given by:

\begin{equation}
\mathbf{U}_k = \{\mathbf{u}_{k,j}\}_{j=1}^{V_k},
\end{equation}

where $\mathbf{u}_{k,j}$ is the prototype for cluster $j$ in class $k$.

\noindent \textbf{Meta-knowledge Compensate Matching.} 
To enhance the clarity of decision boundaries, we ensure that condensed data is similar to its class prototypes and dissimilar to others, optimizing the cosine similarity:

\begin{equation}
\text{sim}(\mathbf{z}_i, \mathbf{u}) = \frac{\mathbf{z}_i \cdot \mathbf{u}}{\|\mathbf{z}_i\| \|\mathbf{u}\| / \tau}
\end{equation}

Here, $\tau$ is a temperature parameter controlling similarity sensitivity. The objective is to contrast current class features with those of other classes, which results in:

\begin{equation}
\mathcal{L}_{\text{MKCL}} = -\log \frac{\sum_{\mathbf{u} \in \mathcal{P}^k} \text{sim}(\mathbf{z}_i, \mathbf{u})}{\sum_{\mathbf{u} \in \mathcal{P}^k} \text{sim}(\mathbf{z}_i, \mathbf{u}) + \sum_{\mathbf{u} \in \mathcal{N}^k} \text{sim}(\mathbf{z}_i, \mathbf{u})}
\end{equation}

Finally, the total objective for exemplar condensation is:

\begin{equation}
\mathcal{L}_{\text{mem}} = \mathcal{L}_{\text{cond}} + \mathcal{L}_{\text{rel}} + \beta \mathcal{L}_{\text{MKCL}},
\end{equation}

where $\beta$ is the weight for meta-knowledge contrastive learning.

\subsection{Prior Knowledge Supervision.}
A key component of the dual-distillation structure is knowledge distillation, which transfers knowledge from previous tasks to mitigate catastrophic forgetting. We apply Knowledge Distillation~\cite{rebuffi2017icarl,li2017learning,wu2019large}, using the soft output of a previously trained teacher model as a regularization term for the current task's student model.

Mathematically, let \( p_{t-1}(x) \) be the teacher model's softmax output after training on task \( t-1 \), and \( p_t(x) \) the student model's output for task \( t \). The objective is to minimize:

\begin{equation}
\mathcal{L}_{\text{KD}} = \mathcal{L}_{\text{ce}} + \lambda \cdot \mathcal{L}_{\text{KL}},
\end{equation}

where \( \mathcal{L}_{\text{ce}} \) is the task-specific cross-entropy loss, and \( \mathcal{L}_{\text{KL}} = \text{KL}(p_{t-1}(x) \parallel p_t(x)) \) is the Kullback-Leibler divergence between the teacher's and student's distributions. The parameter \( \lambda \) balances the task and distillation losses. Minimizing this objective enables the student model to learn the current task while retaining knowledge from prior tasks, mitigating catastrophic forgetting.

\section{Experimental Setup}

\subsection{Implementation details.}
All methods were implemented in PyTorch~\cite{paszke2019pytorch} and executed on an NVIDIA RTX 4090 GPU with an AMD 7950X CPU. We used ResNet18~\cite{he2016deep} as the backbone for feature extraction and FedAvg~\cite{mcmahan2017communication} for global model aggregation. Each task involved $R = 50$ communication rounds, with $E = 30$ local epochs per round. The learning rate was 0.003, and SGD was used as the optimizer. The Elastic Weight Consolidation (EWC) constraint factor was set to 300, the temperature parameter for knowledge distillation was 2, and the distillation loss weight \( \lambda \) was set to 3. Based on grid search, we set \( \beta = 0.5 \) for all experiments.

To simulate non-IID data, we partitioned datasets using the Latent Dirichlet Allocation (LDA) method, adjusting the concentration parameter \( \sigma \) to control data skew. In the CIFAR-100 and TinyImageNet experiments, we started with 20 clients, selecting 10 clients per round, and increased the number of clients by 5 with each new task for 10- and 20-task experiments. For the 50-task CIFAR-100 experiment, we incremented 1 client per task due to data limitations. For the Caltech256 dataset, we started with 5 clients and increased the number of clients by 1 with each new task. And for all experiments, 90\% of existing clients transitioned to new tasks. A detailed breakdown of the data distribution across clients is shown in Figure 1 of \supp{Supplementary Material}.

\subsection{Datasets}
We evaluated the framework on three image classification datasets: CIFAR-100, TinyImageNet, and Caltech-256. \textbf{CIFAR-100}~\cite{krizhevsky2009learning} consists of 60,000 $32 \times 32$ images across 100 classes (600 per class). We used an exemplar memory of 100 per client and tested with 10 tasks (10 classes/task), 20 tasks (5 classes/task), and 50 tasks (2 classes/task). \textbf{TinyImageNet}~\cite{le2015tiny} contains 100,000 $64 \times 64$ images across 200 classes (500 per class), with an exemplar memory of 200 per client and evaluated with 10 tasks (20 classes/task). \textbf{Caltech-256}~\cite{griffin2007caltech} includes 30,607 images across 256 classes (after removing the "background" class). All images were resized to $64 \times 64$ due to computational constraints. We allocated an exemplar memory of 256 per client and tested the method on 16 tasks (16 classes/task).

\subsection{Baselines}
This work compares several baseline methods addressing catastrophic forgetting in federated and class-incremental learning. \textbf{Replay} maintains exemplar memory for replaying prior data. \textbf{iCaRL}\cite{rebuffi2017icarl} integrates representation learning with a nearest-mean-of-exemplars classifier. \textbf{LwF}\cite{li2017learning} preserves prior task knowledge using distillation without accessing old data. \textbf{EWC}\cite{kirkpatrick2017overcoming} uses regularization to retain important weights identified via the Fisher information matrix. \textbf{BiC}\cite{wu2019large} employs bias correction to adjust the decision boundary between old and new classes. \textbf{TARGET}\cite{zhang2023target} introduces exemplar-free distillation leveraging global prototypes for privacy-preserving knowledge retention. \textbf{FedCIL}\cite{qi2023better} combines global distillation with class-balanced sampling to mitigate class imbalance and forgetting. More details are in \supp{Supplementary Material}.

\subsection{Evaluation Metrics}
This work employs several metrics to evaluate federated class-incremental learning. \textbf{Accuracy ($\mathcal{A}$)} measures task-specific performance, including final accuracy ($\mathcal{A}{last}$) and average accuracy ($\mathcal{A}{avg}$). \textbf{Averaged Incremental Accuracy ($\mathcal{A}^{incre}$)}\cite{rebuffi2017icarl} highlights performance throughout incremental learning. \textbf{Accuracy A ($\mathcal{A}^{a}$)}\cite{diaz2018don} balances accuracy across tasks regardless of sample size. \textbf{Backward Transfer (BwT)}\cite{lopez2017gradient} quantifies new task effects on prior tasks, while \textbf{Forward Transfer (FwT)}\cite{lopez2017gradient} evaluates new task benefits for future tasks. \textbf{Remembering}\cite{diaz2018don} measures retention of prior tasks, and \textbf{Forgetting}\cite{chaudhry2018riemannian} assesses information loss across tasks. More details are in \supp{Supplementary Material}.

\section{Results}
\begin{table*}[ht]
  \centering
  \caption{Results on CIFAR100 with 10 tasks and non-IID levels \(\sigma = 0.2, 0.5, 0.8\), evaluated across three metrics: \(\mathcal{A}\), \(\mathcal{A}^{incre}\), and \(\mathcal{A}^{a}\) for the last task and overall performance. \(\Delta\) indicates the absolute difference from ECoral. \\ The ECoral results are highlighted in \capourworkclr{0.2cm}, with improvements in \textcolor{customgreen}{green} and declines in \textcolor{customred}{red}.}
  \label{tab:res_cifar100_10tasks}
  \scalebox{0.96}{
  \begin{tabular}{
    l|
    c
    c
    c
    c|
    c
    c
    c
    c|
    c
    c
    c
    c
    }
    \toprule
    & \multicolumn{4}{c}{$\sigma = 0.2$} & \multicolumn{4}{c}{$\sigma = 0.5$} & \multicolumn{4}{c}{$\sigma = 0.8$} \\
    \cmidrule(lr){2-5} \cmidrule(lr){6-9} \cmidrule(lr){10-13}
    Methods & $\mathcal{A}_{avg}$ & $\Delta$& $\mathcal{A}_{last}$ & $\Delta$& $\mathcal{A}_{avg}$ & $\Delta$& $\mathcal{A}_{last}$ & $\Delta$& $\mathcal{A}_{avg}$ & $\Delta$& $\mathcal{A}_{last}$ & $\Delta$\\
    \midrule
    Replay & {32.82}& {\textcolor{customgreen}{10.70}}& {16.63}& {\textcolor{customgreen}{12.09}}& {36.72} & {\textcolor{customgreen}{12.45}}& {18.72}& {\textcolor{customgreen}{9.25}}& {37.49}& {\textcolor{customgreen}{11.85}}& {18.05}& {\textcolor{customgreen}{13.34}}\\
    iCaRL & {31.97}& {\textcolor{customgreen}{11.55}}& {20.46}& {\textcolor{customgreen}{8.26}}& {43.85} & {\textcolor{customgreen}{5.32}}& {21.76}& {\textcolor{customgreen}{6.21}}& {44.97}& {\textcolor{customgreen}{4.37}}& {23.58}& {\textcolor{customgreen}{7.81}}\\
    EWC & {31.73}& {\textcolor{customgreen}{11.79}}& {14.03}& {\textcolor{customgreen}{14.69}}& {36.55} & {\textcolor{customgreen}{12.62}}& {16.56}& {\textcolor{customgreen}{11.41}}& {38.59}& {\textcolor{customgreen}{10.75}}& {18.42}& {\textcolor{customgreen}{12.97}}\\
    BiC & {28.27}& {\textcolor{customgreen}{15.25}}& {13.08}& {\textcolor{customgreen}{15.64}}& {33.45} & {\textcolor{customgreen}{15.72}}& {17.49}& {\textcolor{customgreen}{10.48}}& {35.62}& {\textcolor{customgreen}{13.72}}& {16.31}& {\textcolor{customgreen}{15.08}}\\
    LwF & {36.64}& {\textcolor{customgreen}{6.88}}& {16.93}& {\textcolor{customgreen}{11.79}}& {43.13} & {\textcolor{customgreen}{6.04}}& {21.38}& {\textcolor{customgreen}{6.59}}& {44.69}& {\textcolor{customgreen}{4.65}}& {22.90}& {\textcolor{customgreen}{8.49}}\\
    TARGET & {30.60}& {\textcolor{customgreen}{12.92}}& {9.42}& {\textcolor{customgreen}{19.30}}& {41.51} & {\textcolor{customgreen}{7.66}}& {16.71}& {\textcolor{customgreen}{11.26}}& {46.05}& {\textcolor{customgreen}{3.29}}& {20.83}& {\textcolor{customgreen}{10.56}}\\
    FedCIL & {30.14}& {\textcolor{customgreen}{13.38}}& {13.62}& {\textcolor{customgreen}{15.10}}& {34.88} & {\textcolor{customgreen}{14.29}}& {15.56}& {\textcolor{customgreen}{12.41}}& {34.98}& {\textcolor{customgreen}{14.36}}& {16.05}& {\textcolor{customgreen}{15.34}}\\
    \rowcolor{ourworkclr!30} ECoral & {43.52}& {--}& {28.72}& {--}& {49.17} & {--}& {27.97}& {--}& {49.34}& {--}& {31.39}& {--}\\
    \midrule
    \midrule
    Methods & $\mathcal{A}_{avg}^{incre}$ & $\Delta$& $\mathcal{A}_{last}^{incre}$& $\Delta$& $\mathcal{A}_{avg}^{incre}$ & $\Delta$& $\mathcal{A}_{last}^{incre}$& $\Delta$& $\mathcal{A}_{avg}^{incre}$ & $\Delta$& $\mathcal{A}_{last}^{incre}$& $\Delta$\\
    \midrule
    Replay & {45.35}& {\textcolor{customgreen}{7.61}}& {36.72}& {\textcolor{customgreen}{12.45}}& {51.01} & {\textcolor{customgreen}{9.48}}& {32.82}& {\textcolor{customgreen}{10.70}}& {52.88}& {\textcolor{customgreen}{7.01}}& {37.49}& {\textcolor{customgreen}{11.85}}\\
    iCaRL & {46.45}& {\textcolor{customgreen}{6.51}}& {43.85}& {\textcolor{customgreen}{5.32}}& {57.37} & {\textcolor{customgreen}{3.12}}& {36.20}& {\textcolor{customgreen}{7.32}}& {57.43}& {\textcolor{customgreen}{2.46}}& {44.97}& {\textcolor{customgreen}{4.37}}\\
    EWC & {46.32}& {\textcolor{customgreen}{6.64}}& {36.55}& {\textcolor{customgreen}{12.62}}& {52.54} & {\textcolor{customgreen}{7.95}}& {31.73}& {\textcolor{customgreen}{11.79}}& {55.35}& {\textcolor{customgreen}{4.54}}& {38.59}& {\textcolor{customgreen}{10.75}}\\
    BiC & {41.57}& {\textcolor{customgreen}{11.39}}& {33.45}& {\textcolor{customgreen}{15.72}}& {48.99} & {\textcolor{customgreen}{11.50}}& {28.27}& {\textcolor{customgreen}{15.25}}& {51.33}& {\textcolor{customgreen}{8.56}}& {35.62}& {\textcolor{customgreen}{13.72}}\\
    LwF & {49.25}& {\textcolor{customgreen}{3.71}}& {43.13}& {\textcolor{customgreen}{6.04}}& {57.18} & {\textcolor{customgreen}{3.31}}& {36.64}& {\textcolor{customgreen}{6.88}}& {59.89}& {--}& {44.69}& {\textcolor{customgreen}{4.65}}\\
    TARGET & {45.50}& {\textcolor{customgreen}{7.46}}& {41.51}& {\textcolor{customgreen}{7.66}}& {57.47} & {\textcolor{customgreen}{3.02}}& {30.60}& {\textcolor{customgreen}{12.92}}& {61.42}& {\textcolor{customred}{1.53}}& {46.05}& {\textcolor{customgreen}{3.29}}\\
    FedCIL & {44.25}& {\textcolor{customgreen}{8.71}}& {34.88}& {\textcolor{customgreen}{14.29}}& {49.84} & {\textcolor{customgreen}{10.65}}& {30.14}& {\textcolor{customgreen}{13.38}}& {50.92}& {\textcolor{customgreen}{8.97}}& {34.98}& {\textcolor{customgreen}{14.36}}\\
    \rowcolor{ourworkclr!30} ECoral & {52.96}& {--}& {49.17}& {--}& {60.49} & {--}& {43.52}& {--}& {59.89}& {--}& {49.34}& {--}\\
    \midrule
    \midrule
    Methods & $\mathcal{A}_{avg}^{a}$ & $\Delta$& $\mathcal{A}_{last}^{a}$& $\Delta$& $\mathcal{A}_{avg}^{a}$ & $\Delta$& $\mathcal{A}_{last}^{a}$& $\Delta$& $\mathcal{A}_{avg}^{a}$ & $\Delta$& $\mathcal{A}_{last}^{a}$& $\Delta$\\
    \midrule
    Replay & {40.39}& {\textcolor{customgreen}{9.19}}& {28.23}& {\textcolor{customgreen}{13.74}}& {45.22} & {\textcolor{customgreen}{11.16}}& {25.26}& {\textcolor{customgreen}{12.20}}& {46.66}& {\textcolor{customgreen}{9.47}} & {28.31}& {\textcolor{customgreen}{14.25}} \\
    iCaRL & {42.68}& {\textcolor{customgreen}{6.90}}& {35.33}& {\textcolor{customgreen}{6.64}}& {52.37} & {\textcolor{customgreen}{4.01}}& {29.71}& {\textcolor{customgreen}{7.75}} & {52.97}& {\textcolor{customgreen}{3.16}} & {36.97}& {\textcolor{customgreen}{5.59}} \\
    EWC & {40.27}& {\textcolor{customgreen}{9.31}}& {27.02}& {\textcolor{customgreen}{14.95}}& {46.09} & {\textcolor{customgreen}{10.29}}& {23.19}& {\textcolor{customgreen}{14.27}} & {48.53}& {\textcolor{customgreen}{7.60}} & {28.65}& {\textcolor{customgreen}{13.91}} \\
    BiC & {36.02}& {\textcolor{customgreen}{13.56}}& {24.46}& {\textcolor{customgreen}{17.51}}& {42.44} & {\textcolor{customgreen}{13.94}}& {20.52}& {\textcolor{customgreen}{16.94}} & {44.87}& {\textcolor{customgreen}{11.26}} & {26.36}& {\textcolor{customgreen}{16.20}} \\
    LwF & {44.25}& {\textcolor{customgreen}{5.33}}& {34.40}& {\textcolor{customgreen}{7.57}}& {51.85} & {\textcolor{customgreen}{4.53}}& {29.03}& {\textcolor{customgreen}{8.43}} & {54.06}& {\textcolor{customgreen}{2.07}} & {35.32}& {\textcolor{customgreen}{7.24}} \\
    TARGET & {39.65}& {\textcolor{customgreen}{9.93}}& {31.13}& {\textcolor{customgreen}{10.84}}& {51.88} & {\textcolor{customgreen}{4.50}}& {21.54}& {\textcolor{customgreen}{15.92}} & {55.98}& {\textcolor{customgreen}{0.15}} & {36.12}& {\textcolor{customgreen}{6.44}} \\
    FedCIL & {38.43}& {\textcolor{customgreen}{11.15}}& {25.84}& {\textcolor{customgreen}{16.13}}& {43.91} & {\textcolor{customgreen}{12.47}} & {21.84}& {\textcolor{customgreen}{15.62}} & {44.35}& {\textcolor{customgreen}{11.78}} & {25.60}& {\textcolor{customgreen}{16.96}} \\
    \rowcolor{ourworkclr!30} ECoral & {49.58}& {--}& {41.97}& {--}& {56.38} & {--}& {37.46}& {--}& {56.13}& {--}& {42.56}& {--}\\
    \bottomrule
    \end{tabular}
    }
\end{table*}

\noindent \textbf{ECoral efficiently mitigates forgetting.}
As shown in Table~\ref{tab:res_cifar100_10tasks} ($\sigma=0.5$) and Table~\ref{tab:res_cal_tiny_10tasks} with more complex datasets, ECoral consistently outperforms baseline methods. On CIFAR-100 at $\sigma=0.5$, ECoral achieves an average accuracy ($\mathcal{A}{avg}$) of 49.17\% and last-task accuracy ($\mathcal{A}{last}$) of 27.97\%, surpassing iCaRL, which has an $\mathcal{A}{avg}$ of 43.85\% and $\mathcal{A}{last}$ of 21.76\%. This highlights ECoral’s ability to mitigate catastrophic forgetting and maintain strong performance across tasks.

For more complex datasets, such as Tiny-ImageNet and Caltech-256, ECoral remains effective. On Tiny-ImageNet, ECoral achieves $\mathcal{A}{avg}$ of 38.78\%, outperforming iCaRL’s 36.46\%. While iCaRL slightly surpasses ECoral in last-task accuracy (22.80\% vs. 20.88\%), ECoral maintains a better overall balance across tasks. A similar trend is observed with Caltech-256, where ECoral achieves $\mathcal{A}{avg}$ of 31.06\%, though iCaRL leads in last-task accuracy (23.52\% vs. 21.66\%).

While iCaRL slightly outperforms ECoral in last-task accuracy in some cases, ECoral excels in overall task performance, demonstrating its strength in mitigating forgetting and maintaining balanced performance across tasks.

\begin{table*}
  \centering
  \caption{Results on Tiny-ImageNet with 10 tasks (10 classes per task) and on Caltech-256 with 16 tasks (16 classes per task), both under a Non-IID setting with $\sigma=0.5$. \\ The results row of our ECoral is highlighted in \capourworkclr{0.2cm}. The best result is highlighted with \capbestresult{0.2cm}, and the second-best result is highlighted with \capsecondbestresult{0.2cm}.}
  \label{tab:res_cal_tiny_10tasks}
  \scalebox{1}{
  \begin{tabular}{l|cccccc|cccccc}
    \toprule
    & \multicolumn{6}{c}{Tiny-Imagenet (10 Tasks)} & \multicolumn{6}{c}{Caltech-256 (16 Tasks)} \\
    \cmidrule(lr){2-7} \cmidrule(lr){8-13}
    Methods & $\mathcal{A}_{avg}$ & $\mathcal{A}_{last}$ & $\mathcal{A}_{avg}^{incre}$ & $\mathcal{A}_{last}^{incre}$ & $\mathcal{A}_{avg}^{a}$ & $\mathcal{A}_{last}^{a}$ & $\mathcal{A}_{avg}$ & $\mathcal{A}_{last}$ & $\mathcal{A}_{avg}^{incre}$ & $\mathcal{A}_{last}^{incre}$ & $\mathcal{A}_{avg}^{a}$ & $\mathcal{A}_{last}^{a}$ \\
    \midrule
    Replay & {35.73}& {18.73}& {46.04}& {33.51}& {41.72}& {26.17}& {24.24}& {20.92}& {31.46}& {24.24}& {28.20}& {20.79}\\
    iCaRL & {\cellcolor{secondbestresult} 36.46}& {\cellcolor{bestresult} 22.80}& {44.92}& {\cellcolor{secondbestresult} 35.40}& {41.56}& {\cellcolor{secondbestresult} 29.69}& {26.72}& \cellcolor{bestresult}{23.52}& {33.75}& {26.72}& {30.54}& {23.11}\\
    EWC & {31.10}& {13.10}& {45.23}& {30.14}& {39.19}& {21.49}& {20.68}& {11.08}& {34.17}& {20.68}& {28.12}& {13.67}\\
    BiC & {35.88}& {20.46}& {46.31}& {34.51}& {42.07}& {27.57}& {\cellcolor{secondbestresult}29.26}& {\cellcolor{secondbestresult}22.70}& {\cellcolor{secondbestresult}37.78}& {\cellcolor{secondbestresult}29.26}& {\cellcolor{secondbestresult}33.84}& {\cellcolor{secondbestresult}24.51}\\
    LwF & {35.76}& {16.78}& {\cellcolor{secondbestresult} 47.51}& {33.64}& {\cellcolor{secondbestresult} 42.53}& {25.54}& {23.20}& {12.88}& {35.51}& {23.20}& {30.06}& {16.67}\\
    TARGET & {27.00}& {10.49}& {40.83}& {25.46}& {34.63}& {16.90}& {20.46}& {10.31}& {35.28}& {20.46}& {27.83}& {12.15}\\
    FedCIL & {30.18}& {12.59}& {44.67}& {29.11}& {38.43}& {20.13}& {18.53}& {10.53}& {31.16}& {18.53}& {25.59}& {11.92}\\
    \rowcolor{ourworkclr!30} ECoral & {\cellcolor{bestresult} 38.78}& {\cellcolor{secondbestresult} 20.88}& {\cellcolor{bestresult} 48.46}& {\cellcolor{bestresult} 37.80}& {\cellcolor{bestresult} 44.94}& {\cellcolor{bestresult} 31.24}& {\cellcolor{bestresult}31.06}& {21.66}& {\cellcolor{bestresult}40.64}& {\cellcolor{bestresult}31.06}& {\cellcolor{bestresult}36.23}& {\cellcolor{bestresult} 25.11}\\
    \bottomrule
  \end{tabular}
  }
\end{table*}

\noindent \textbf{ECoral is keep effective at different levels of non-iid.}
As shown in Table~\ref{tab:res_cifar100_10tasks}, ECoral outperforms baseline methods in three key metrics—Accuracy ($\mathcal{A}$), Averaged Incremental Accuracy ($\mathcal{A}^{incre}$), and Accuracy A ($\mathcal{A}^{a}$)—across various non-IID distributions. ECoral consistently achieves higher final accuracy ($\mathcal{A}{last}$) and average accuracy ($\mathcal{A}{avg}$), particularly excelling in challenging non-IID settings, such as $\sigma = 0.2$, the most skewed scenario. In this setting, ECoral effectively retains knowledge from previous tasks and adapts to new ones, significantly reducing catastrophic forgetting.

At $\sigma = 0.2$, ECoral shows significant improvements in both $\mathcal{A}{last}$ and $\mathcal{A}{avg}$ compared to other methods. It also maintains superior Averaged Incremental Accuracy ($\mathcal{A}^{incre}$) throughout the learning process. As non-IID severity decreases (e.g., $\sigma = 0.5$, $\sigma = 0.8$), ECoral continues to outperform other approaches, demonstrating adaptability across different data skews. ECoral also excels in $\mathcal{A}^{a}$, balancing performance across tasks regardless of sample size. In the $\sigma = 0.2$ scenario, ECoral achieves $\mathcal{A}^{a}_{avg}$ of 49.58\%, outperforming the next best method by a significant margin.

These results highlight ECoral’s robustness in mitigating catastrophic forgetting and improving task performance, especially in highly non-IID environments.

\noindent \textbf{ECoral achieves superior performance across multiple evaluation metrics.}
As shown in Figure~\ref{fig:all_metrics_cifar100_t10}, ECoral excels in key metrics such as BwT, FwT, Forgetting, and Remembering. It shows lower negative BwT compared to several baselines, effectively limiting the detrimental effects on previously learned tasks. While iCaRL and Target have slightly better BwT early on, ECoral avoids the significant performance decline seen in methods like FedCIL and BiC, demonstrating better retention of prior knowledge.

ECoral also displays strong FwT, with earlier tasks contributing positively to the performance on future tasks. In non-IID settings, where task distributions vary, ECoral leverages prior knowledge to improve performance on new tasks, outperforming FedCIL and BiC, which struggle in this regard. Additionally, ECoral exhibits lower forgetting, especially in later task stages, indicating better long-term knowledge retention and resistance to catastrophic forgetting compared to methods like FedCIL and BiC.
\begin{figure}[ht]
  \centering
  \includegraphics[width=\linewidth]{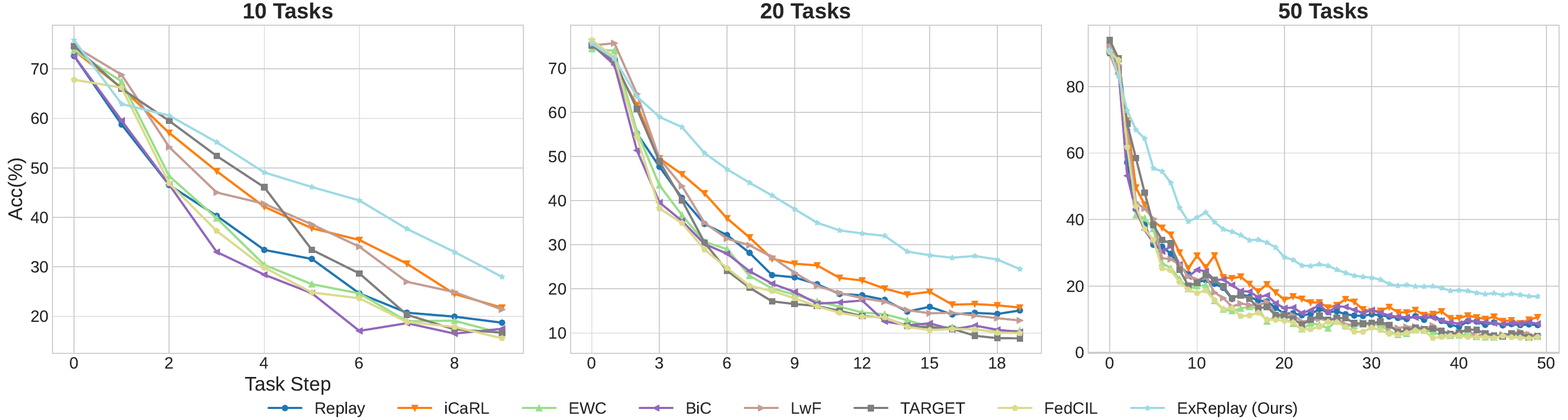}
  \caption{Performance evaluation on CIFAR100 under a Non-IID setting with $\sigma=0.5$. The final accuracy $\mathcal{A}$ (\%) is reported after learning each task. The left plot shows results with 10 steps (10 classes per task), the middle with 20 steps (5 classes per task), and the right with 50 steps (2 classes per task).}
  \label{fig:all_acc_cifar100}
\end{figure}
In terms of Remembering, ECoral performs competitively, retaining knowledge effectively. While iCaRL slightly outperforms ECoral in some cases, ECoral strikes a strong balance between retention, forward transfer, and reduced forgetting, ensuring robust long-term performance.

\begin{figure}[ht]
  \centering
  \includegraphics[width=\linewidth]{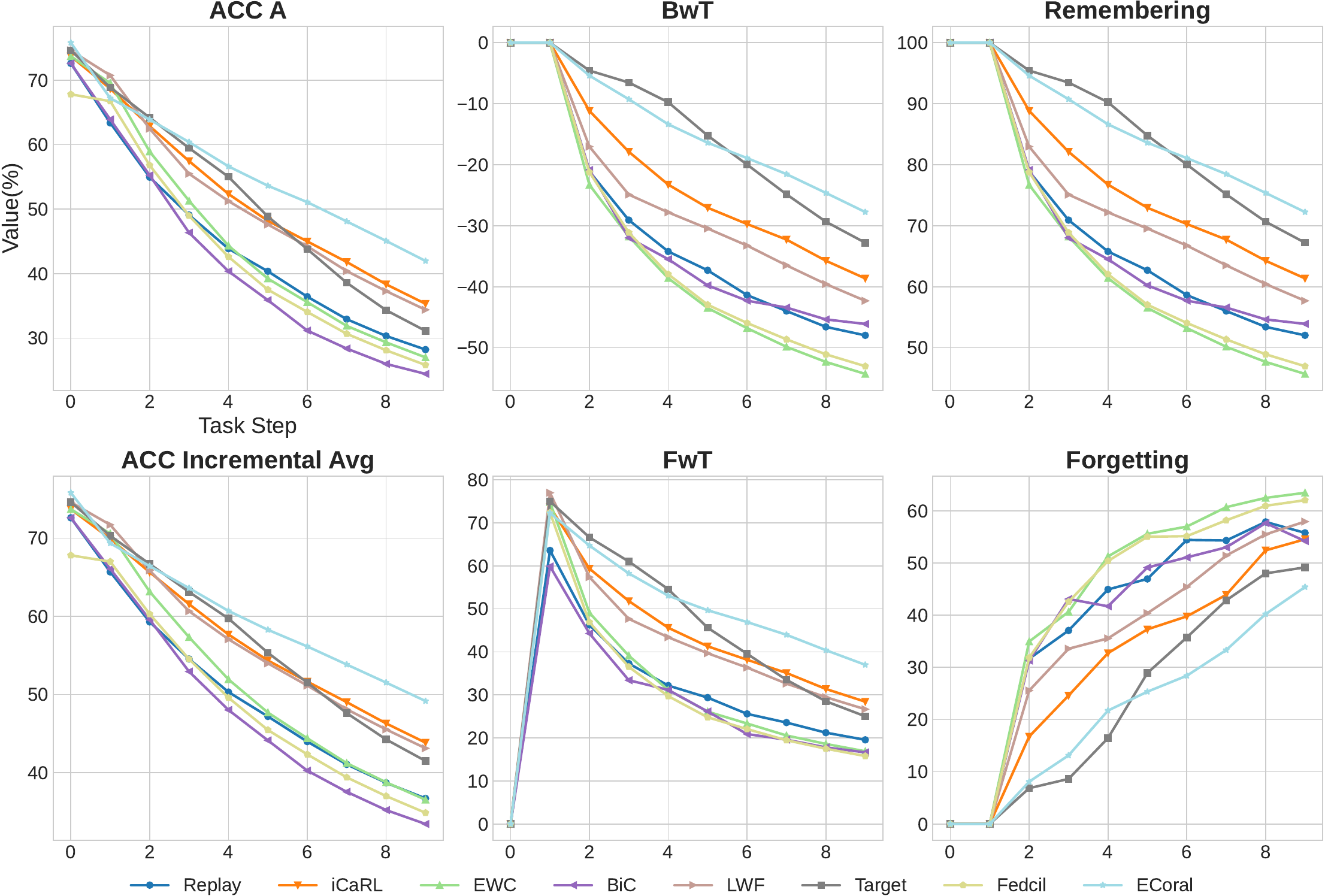}
  \caption{Evaluation of multiple metrics (\%) on CIFAR100 under a Non-IID setting with $\sigma=0.5$, across a total of 10 tasks.}
  \label{fig:all_metrics_cifar100_t10}
\end{figure}

\noindent \textbf{ECoral can perform consistently in a long-term training task.}
As shown in Figure~\ref{fig:all_acc_cifar100}, ECoral significantly outperforms baseline methods in both 20-task and 50-task continual learning setups, especially in average and final accuracy. In the 20-task setup, ECoral achieves an average accuracy of 63.60\%, surpassing BiC (51.4\%) and Replay (55.27\%). By the final task, ECoral retains 59.00\% accuracy, while BiC (39.6\%) and FedCIL (38.25\%) experience significant declines.

In the more challenging 50-task setup, ECoral maintains a strong performance with an average accuracy of 91.00\%, outperforming BiC (90.50\%) and iCaRL (90.00\%). By the final task, ECoral achieves 64.40\% accuracy, far exceeding BiC (36.90\%) and Replay (38.40\%). These results underscore ECoral’s ability to retain knowledge and perform consistently in both short- and long-term continual learning, demonstrating its robustness and scalability in federated learning.

\begin{figure}[ht]
  \centering
  \includegraphics[width=0.8\linewidth]{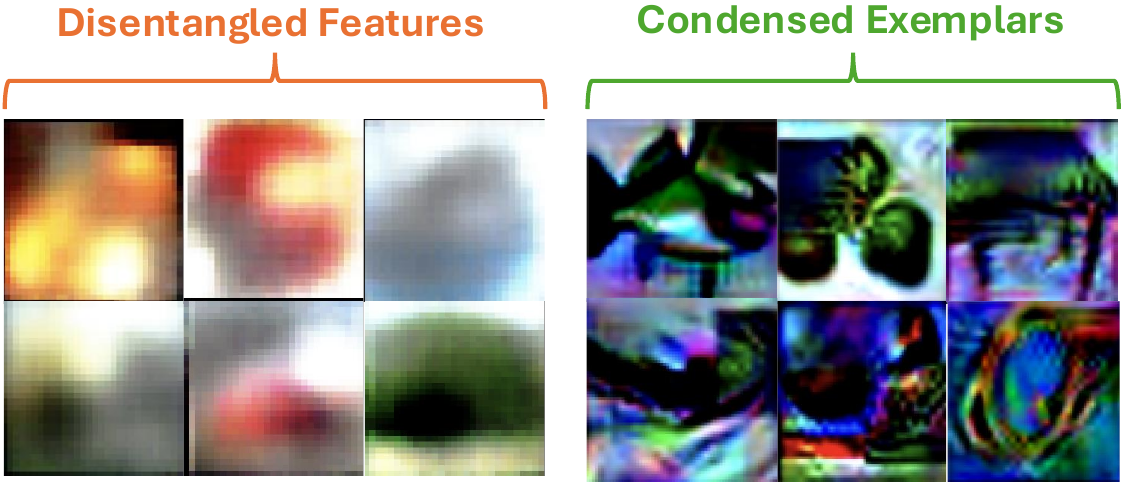}
  \caption{Examples of disentangled features from Shared-VAE and final condensed exemplars.}
  \label{fig:recon_conden_data}
\end{figure}

\begin{table}[t]
\centering
    \scalebox{0.86}{
    \begin{tabular}{c|c|c|c|c|c|c|c|c|c}
    \toprule
    \multicolumn{1}{c|}{\textbf{Method}} & \textbf{A} & \textbf{G} & \textbf{F} & \textbf{C} & \textbf{K} & \textbf{Avg} & $\Delta$ & \textbf{Last} & $\Delta$ \\
    \midrule
    Replay &  &  &  &  &  & 36.72 & - & 18.72 & - \\
    \cmidrule{1-10}
    & \checkmark &  &  & & &  38.12&  1.40&  20.12& 1.40\\
    \cmidrule{2-10}
    & \checkmark & \checkmark & & & & 40.25& 3.53& 21.56& 2.84\\
    \cmidrule{2-10}
    ECoral & \checkmark & \checkmark & \checkmark & & & 40.92& 4.20& 22.33& 3.61\\
    \cmidrule{2-10}
    & \checkmark & \checkmark & \checkmark & \checkmark & & 41.73& 5.01& 24.14& 5.42\\
    \cmidrule{2-10}
    & \checkmark & \checkmark & \checkmark & \checkmark & \checkmark & 49.17 & 12.45 & 27.97 & 9.25\\
    \bottomrule
    \end{tabular}
}
\caption{Ablation study of ECoral on CIFAR100 with 10 tasks and non-IID level \( \sigma = 0.5 \). Improvement compared to the Replay baseline is marked as \( \Delta \). Components: A (Adjustable Memory), G (Gradient Matching), F (Feature Matching), C (Compensate Matching).}
\vspace{-0.3cm}
\label{table:ab}
\end{table}

\noindent \textbf{ECoral is user privacy friendly.}
This work addresses user privacy in two ways. First, the Shared-VAE model prevents the regeneration of raw data from other clients, and the generated data remains semantically uninterpretable, ensuring privacy at the federated learning (FL) level. Second, condensed data is designed to be identifiable only to the client’s local classes and indecipherable by humans, protecting privacy during memory replay.

Figure~\ref{fig:recon_conden_data} shows six disentangled features generated by Shared-VAE and six condensed exemplars. The disentangled features (left panel) are abstract, containing only basic information like colors and vague outlines, making them unreadable to humans. In the right panel, the first row of condensed exemplars is loosely linked to categories but lacks detail, ensuring no specific client data is exposed. The second row is completely unrecognizable, further enhancing data privacy. Importantly, these exemplars are derived only from the client’s local data, ensuring no sensitive information from other clients is included.



\subsection{Ablation Study}

The ablation study in Table~\ref{table:ab} shows the impact of each component in ECoral. Adding Adjustable Memory gives a modest improvement, emphasizing the importance of efficient memory allocation. Gradient Matching (+3.53\%) helps align exemplars with new data, improving task transfer. Feature Matching (+4.20\%) ensures consistency between real images and exemplars, while Compensation Matching (+5.01\%) addresses meta-knowledge heterogeneity, crucial for non-IID data. The full ECoral method yields the best performance (+12.45\% average, +9.25\% on the last task), demonstrating that each component contributes to mitigating catastrophic forgetting and improving performance across tasks.

\section{Conclusion}
In this paper, we propose the ECoral framework, which successfully addresses critical challenges in Federated Class-Incremental Learning (FCIL) by enhancing memory efficiency and improving resilience against catastrophic forgetting. The combination of exemplar condensation and meta-knowledge contrastive learning allows the model to store more informative and privacy-preserving condensed exemplars, while client-wise feature disentanglement mitigates the negative effects of data heterogeneity. This approach ensures consistent performance across highly non-IID environments, making it well-suited for real-world federated learning applications where data privacy and resource constraints are key concerns. However, we observe that ECoral’s advantage decreases when applied to more complex datasets. Future work will focus on strengthening ECoral’s performance in these complex scenarios, ensuring robustness and scalability across diverse real-world data challenges.

\bibliographystyle{named}
\bibliography{my.bib}

\clearpage
\appendix
\section{Full Preliminaries}
\label{sec:app_pre}
\noindent{\textbf{Federated Class-incremental Learning.}}
Federated Class-incremental Learning (FCIL) aims to collaboratively train a global model using streaming data that sequentially introduces new classes. In this context, a model training process consists of a series of sequential tasks $\mathcal{T} = \{\mathcal{T}^t\}_{t=1}^T$, where $T$ denotes the total number of tasks. The system involves $C$ local clients and a central server $S_g$. Each task comprises $R$ global communication rounds (where $r = 1, \dots, R$), and in each round $r$, a subset of the local participants is randomly selected for gradient aggregation. When the $l$-th client $C_l$ is selected for a given global round in the $t$-th incremental task, it receives the latest global model $\theta^{r,t}$. 

Drawing inspiration from online learning, each client maintains a fixed-size local memory $\mathcal{M}_l = \left\{\left(\mathbf{x}_{l, m}, \mathbf{y}_{l, m}\right)\right\}_{m=1}^M$ of size $M$, storing examples from prior tasks for knowledge replay. In this work, we divide this memory into three parts: $\mathcal{M}_{\text{orig}}$, which holds original data sampled from the current task’s training set; $\mathcal{M}_{\text{cond}}$, which stores condensed exemplars from prior tasks; and $\mathcal{M}_{\text{sum}}$, which saves summarizing data from the current task. At each iteration of the current task, a batch of samples $\mathbf{B}_m = \left\{\left(\mathbf{x}_{i, m}, \mathbf{y}^{i, m}\right)\right\}_{i=1}^{B_m}$ is randomly drawn from the memory and jointly trained alongside the current task data $\mathbf{B}_n = \left\{\left(\mathbf{x}^{t}_{i}, \mathbf{y}^{t}_{i}\right)\right\}_{i=1}^{B_n}$. Here, $B_m \leq M$ and $B_n$ represent the mini-batch sizes of the replayed data and the current task data, respectively. The joint training objective is expressed as:
\begin{equation}
    \theta^{r,t}_l = \arg\min_{\theta^{r,t}} \mathcal{L}(\theta^{r,t}; \mathbf{B}_n) + \lambda \mathcal{L}_m(\theta^{r,t}; \mathbf{B}_m),
\end{equation}
where $\mathcal{L}$ and $\mathcal{L}_m$ are the loss functions for the current task data and memory data, respectively. $\lambda$ is a hyper-parameter for regulation.

The client trains the global model $\theta^{r,t}$ on its own $t$-th incremental task data $\mathcal{D}_l^t \cup \mathcal{M}_l$, where $\mathcal{D}_l^t = \left\{\left(\mathbf{x}_{l, i}^t, \mathbf{y}_{l, i}^t\right)\right\}_{i=1}^{N_l^t} \subset \mathcal{T}^t$ represents the training data for new categories specific to the $l$-th client. The category distribution for the $l$-th client is denoted by $\mathbf{P}_l$. The distributions $\{\mathbf{P}_l\}_{l=1}^C$ are non-independent and identically distributed (non-IID). At the $t$-th incremental task, the label space $\mathcal{Y}_l^t \subseteq \mathcal{Y}^t$ for the $l$-th local client is a subset of $\mathcal{Y}^t = \bigcup_{l=1}^C \mathcal{Y}_l^t$, which includes $\mathcal{K}_l^t$ new categories ($\mathcal{K}_l^t \leq \mathcal{K}^t$), distinct from the previous categories $\mathcal{K}_l^p = \sum_{i=1}^{t-1} \mathcal{K}_l^i \subseteq \bigcup_{j=1}^{t-1} \mathcal{Y}_l^j$. After receiving $\theta^{r,t}$ and performing local training on the $t$-th incremental task, the $l$-th client obtains an updated model $\theta^{r,t}_l$. These locally updated models from selected clients are then uploaded to the central server $S_g$, where they are aggregated to form the new global model $\theta^{r+1,t}$ for the next round. The central server $S_g$ subsequently distributes the updated parameters $\theta^{r+1,t}$ to the local clients for the following global round.

\noindent{\textbf{Client increment strategy.}}
To better simulate a real-world federated continual learning application, we adopt the client increment strategy introduced in GLFC \cite{dong2022federated}. This strategy divides local participants into three dynamic groups for each incremental task: Old ($\mathcal{G}_o$), In-between ($\mathcal{G}_b$), and New ($\mathcal{G}_n$). The Old group ($\mathcal{G}_o$), consisting of $G_o$ participants, only has access to data from classes introduced in previous tasks and does not receive any data for the new task. The In-between group ($\mathcal{G}_b$), with $G_b$ members, works with both the new classes from the current task and the classes from the previous task. Finally, the New group ($\mathcal{G}_n$), comprising $G_n$ newly added participants, focuses exclusively on data containing new classes from the current task.

The group compositions are dynamically updated with the progression of tasks. Specifically, the membership of the groups ${\mathcal{G}_o, \mathcal{G}_b, \mathcal{G}_n}$ is redefined randomly at each global round, and new participants are irregularly added to $\mathcal{G}_n$ as incremental tasks arrive. This incremental process gradually increases the total number of participants, $G = G_o + G_b + G_n$, as more tasks are introduced, closely mimicking the nature of streaming data in real-world FL applications.

\subsection{Problem Definition}

\subsubsection*{Forgetting in FCIL}
The primary objective of global model optimization at the $t$-th incremental task is to minimize the classification error across the current category set $\mathcal{K}_t$. However, when a new task arises, clients are often constrained by privacy restrictions and limited resources, allowing only restricted access to data from previous tasks. The category imbalance between old and new categories ($\mathcal{T}_l^t$ and $\mathcal{M}_l$) at the local level exacerbates this issue, leading to significant performance degradation during local training. This limitation frequently results in a notable decline in performance on earlier tasks, a phenomenon known as catastrophic forgetting. To mitigate catastrophic forgetting in the global model, our goal is to minimize the classification error on the current category set $\mathcal{K}_t$ while simultaneously preserving the knowledge of previously learned categories. The objective function is formally defined as:
\begin{equation}
\min _{\theta_t} \sum_{k \in \mathcal{K}_t} \sum_{i=1}^{N_k} \mathcal{L}\left(\mathbf{P}_l^t\left(\mathbf{x}_{l,i}^t ; \theta_{r,t}\right), \mathbf{y}_{l,i}^t\right)
\end{equation}
where $\mathcal{L}$ is a loss function that measures the classification error, and $N_k$ is the number of samples in class $k$.

\subsubsection*{Meta-information Heterogeneity} \label{sec:meta-info-hetero}
The condensation of data from non-IID sources inherently retains the non-IID characteristics at an information level, leading to what we define as the meta-information heterogeneity problem. Given that each client's original dataset \( \mathcal{D}^t_l \) on task $t$-th is drawn from a unique distribution \( \mathbf{P}_l(X, Y) \), the resulting condensed exemplar dataset \( \mathcal{M}_{\text{cond}} \), optimized to represent \( \mathcal{D}^t_l \), will also reflect this distinct distribution. Mathematically, this is expressed as \( \mathbf{P}_l^{\text{cond}}(X, Y) \neq \mathbf{P}_{l'}^{\text{cond}}(X, Y) \) for some clients \( l \neq l' \). The divergence in information content between these condensed datasets can be quantified using measures such as Kullback-Leibler (KL) divergence, where \( \text{KL}(\mathcal{I}(\mathcal{T}_l^{\text{cond}}) \parallel \mathcal{I}(\mathcal{T}_{l'}^{\text{cond}})) > 0 \) indicates non-identical information content across clients, thus confirming meta-information heterogeneity. Non-IID data has been shown to exacerbate catastrophic forgetting, as explored in \cite{zhang2023target}, further complicating federated continual learning. Similarly, when these heterogeneous condensed datasets are used to train a global model \( \theta^{r,t} \), the model’s updates from different clients may conflict due to the diverse information content, leading to suboptimal performance. This degradation is reflected in the global loss function \( \mathcal{L}(\theta) \), which generally increases compared to an IID scenario, expressed as \( \Delta \mathcal{L} = \mathcal{L}_{\text{non-iid}}(\theta^{r,t}) - \mathcal{L}_{\text{iid}}(\theta^{r,t}) > 0 \). Therefore, condensed datasets from non-IID sources introduce a meta-information heterogeneity problem that adversely affects the global model’s performance, mirroring the challenges posed by non-IID data in traditional federated learning.

\section{More Experiments Details}

\subsection{Data Distribution}
\begin{figure}[ht]
  \centering
  \includegraphics[width=\linewidth]{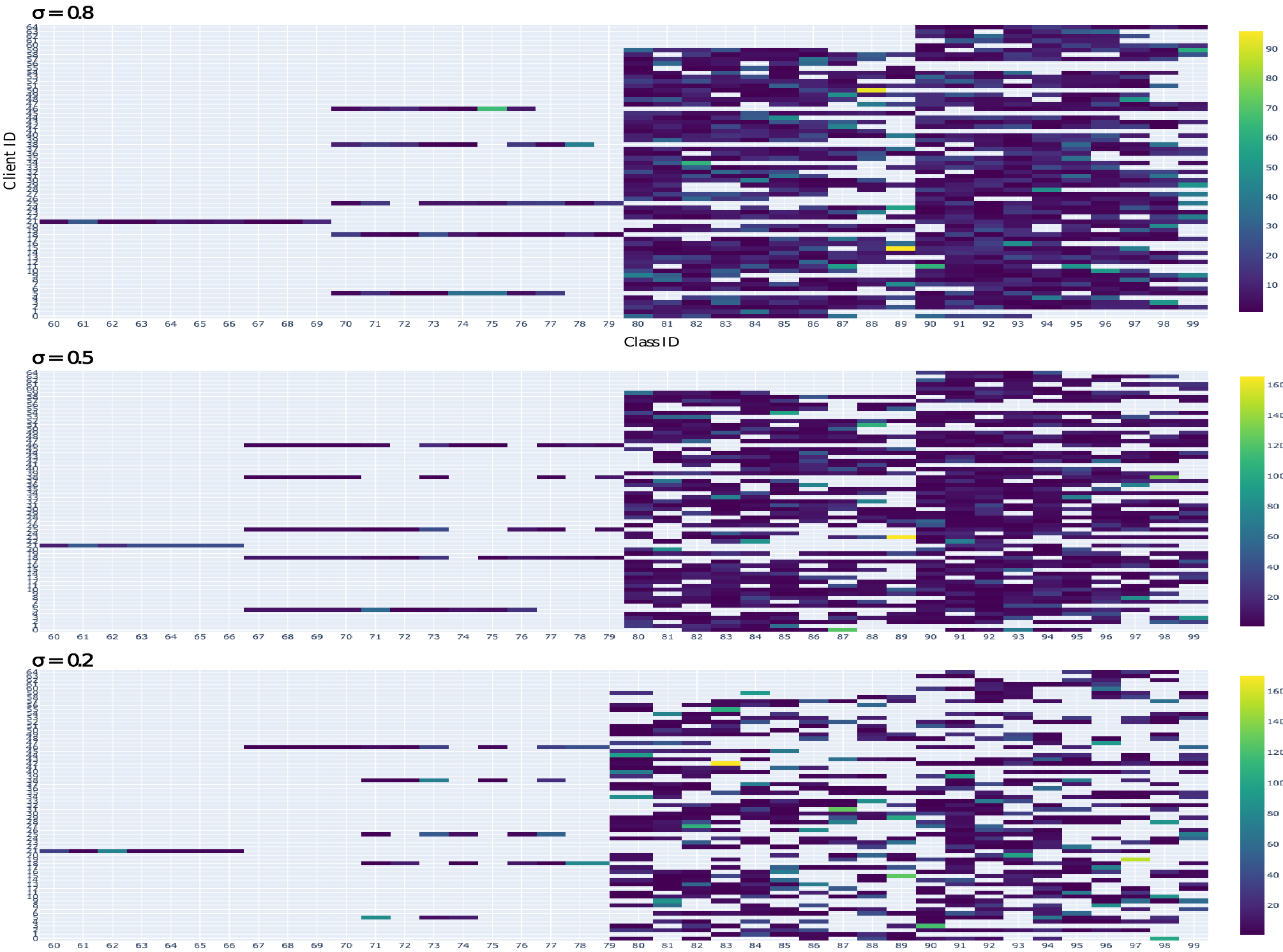}
  \caption{Training data distribution of every client for CIFAR-100 on the final task, with non-IID levels $\sigma$ of 0.2, 0.5, and 0.8 across a total of 10 tasks (each task containing 10 classes).}
  \label{fig:data_distri}
\end{figure}

To clearly illustrate the data distributions across clients under varying degrees of non-IID settings (controlled by $\sigma$), Figure~\ref{fig:data_distri} presents the data distribution for the final task in the CIFAR-100 dataset experiment with a total of 10 tasks.
\label{sec:app_data_dis}

\subsection{Baselines Details}
\label{sec:app_baselines}
\noindent \textbf{Replay} maintains an exemplar memory at each client to store and replay a subset of previous data, mitigating catastrophic forgetting in federated learning settings by randomly selecting samples from the training data and incrementally adding new classes with each new task.

\noindent \textbf{iCaRL}~\cite{rebuffi2017icarl} proposes an incremental learning method that integrates representation learning with a nearest-mean-of-exemplars classifier, utilizing a fixed memory budget to store exemplars from previous classes, thereby mitigating catastrophic forgetting while learning new classes.

\noindent \textbf{LwF}~\cite{li2017learning} enables a neural network to learn new tasks without forgetting previously learned tasks by using knowledge distillation to preserve the model’s responses on old tasks during training, all without requiring access to the original data from the old tasks.

\noindent \textbf{EWC}~\cite{kirkpatrick2017overcoming} mitigates catastrophic forgetting in neural networks by adding a regularization term that penalizes changes to important weights, identified using the Fisher information matrix. This approach allows the model to learn new tasks while preserving performance on previously learned tasks without requiring access to old data.

\noindent \textbf{BiC}~\cite{wu2019large} tackles the bias toward new classes in class-incremental learning by introducing a two-stage training framework that adds a bias correction layer, which is fine-tuned using a small validation set to adjust the decision boundary between old and new classes, effectively reducing bias and improving classification accuracy.

\noindent \textbf{TARGET}~\cite{zhang2023target} addresses federated class-continual learning by introducing an exemplar-free distillation method that utilizes global prototypes to preserve knowledge of previous classes without storing or generating data, effectively mitigating catastrophic forgetting in a privacy-preserving manner.

\noindent \textbf{FedCIL}~\cite{qi2023better} addresses federated class-incremental learning by introducing a global knowledge distillation method to preserve knowledge of old classes and a class-balanced sampling strategy to mitigate class imbalance, enabling clients to learn new classes while reducing catastrophic forgetting incrementally.

\subsection{Evaluation Metrics Details}
\begin{figure*}[ht]
  \centering
  \includegraphics[width=\linewidth]{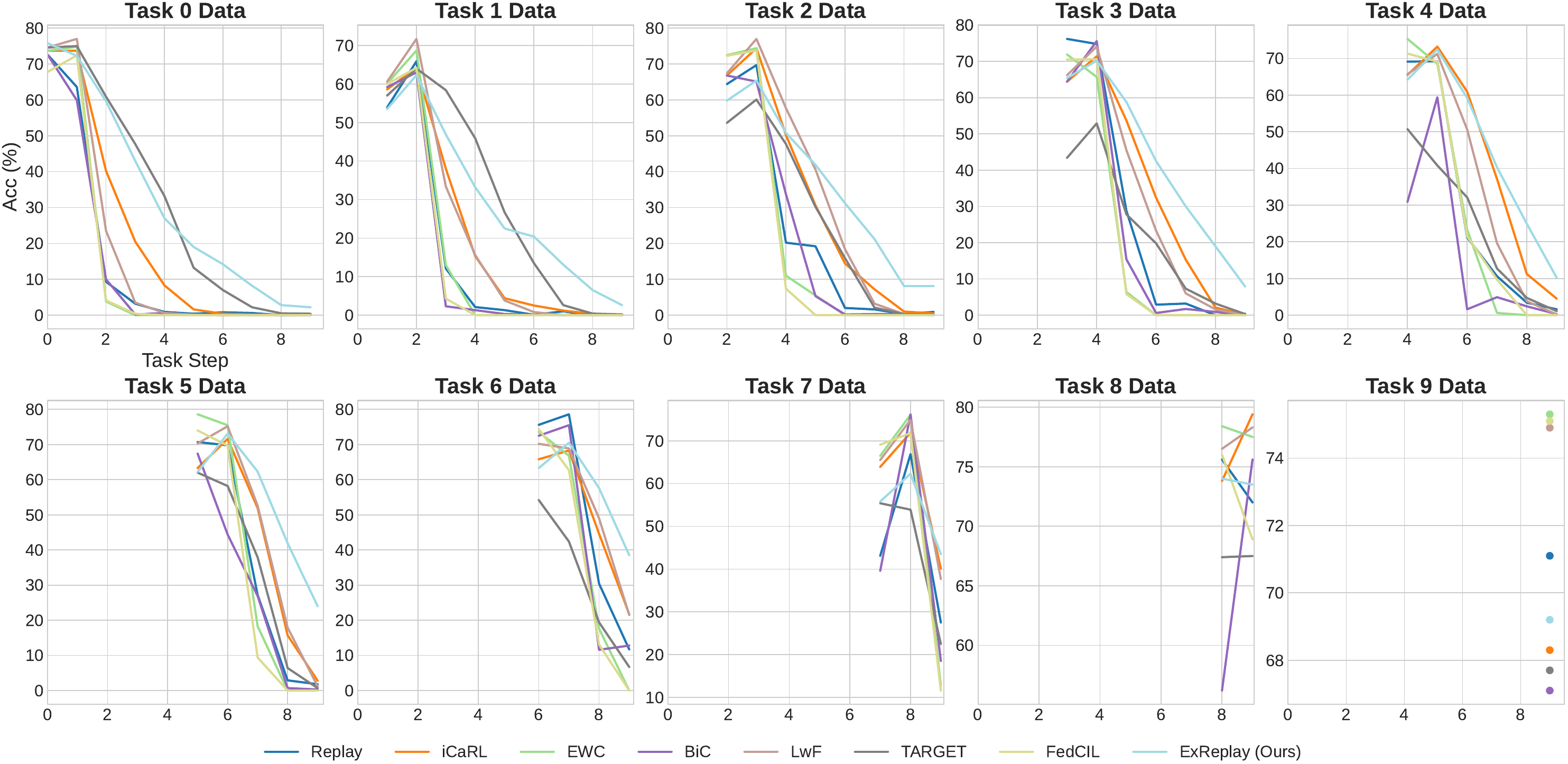}
  \caption{Performance evaluation on CIFAR100 under a Non-IID setting with $\sigma=0.5$, across 10 tasks. The final accuracy $\mathcal{A}$ (\%) for each learned task is reported after the completion of each task.}
  \label{fig:cifar100_task10_per_task}
\end{figure*}

\noindent \textbf{Accuracy ($\mathcal{A}$)}: This metric computes the accuracy for a given task. We report the final accuracy after all tasks have been trained as $\mathcal{A}_{last}$, and the average accuracy across the last round of every task as $\mathcal{A}_{avg}$.

\noindent \textbf{Averaged Incremental Accuracy $\mathcal{A}^{incre}$}~\cite{rebuffi2017icarl}: This metric calculates the average accuracy after the completion of each task, emphasizing the model’s performance throughout the incremental learning process. We denote the overall averaged accuracy across all tasks as $\mathcal{A}_{avg}^{incre}$, and the accuracy after the last task as $\mathcal{A}_{last}^{incre}$.


\noindent \textbf{Accuracy A ($\mathcal{A}^{a}$)}~\cite{diaz2018don}: As defined by Díaz-Rodríguez et al., this metric differs from standard accuracy by assigning equal weight to the accuracy of each task, regardless of the number of samples. For instance, in a scenario where Task 1 has 50,000 images and Task 2 has 1,000 images, standard accuracy would give more weight to Task 1, whereas Accuracy A treats both tasks equally. We denote the overall averaged Accuracy A as $\mathcal{A}_{avg}^{a}$ and the Accuracy A after the last task as $\mathcal{A}_{last}^{a}$.

\noindent \textbf{Backward Transfer (BwT)}~\cite{lopez2017gradient}: This metric measures the influence that learning a new task has on the performance of previously learned tasks. A positive BwT indicates an improvement in past tasks after learning new ones, while a negative BwT signifies forgetting. It is denoted as BwT.


\noindent \textbf{Forward Transfer (FwT)}~\cite{lopez2017gradient}: This metric assesses the influence that learning a new task has on the performance of future tasks. Positive forward transfer implies that learning prior tasks benefits future tasks, enhancing initial performance. It is denoted as FwT.


\noindent \textbf{Remembering}~\cite{diaz2018don}: This metric calculates the degree of retention for previous tasks as part of the backward transfer process. It quantifies how well the model remembers earlier tasks after learning new ones.



\noindent \textbf{Forgetting}~\cite{chaudhry2018riemannian}: This metric measures the average amount of forgetting across all tasks, helping to quantify how much information is lost as new tasks are learned. It is calculated by comparing the maximum performance on a task with its performance after learning subsequent tasks.



\section{Additional Results}

\noindent \textbf{ECoral balance the knowledge learned in each task.}
As illustrated in Figure~\ref{fig:cifar100_task10_per_task}, the CIFAR-100 experiments with $\sigma=0.5$ provide a detailed analysis of ECoral’s performance as tasks are progressively introduced, reflecting a typical continual learning scenario. In the early stage, for the first task (T0), all methods show strong performance, with TARGET and iCaRL slightly outperforming ECoral in the initial steps. Nonetheless, ECoral remains highly competitive, demonstrating a robust ability to learn and adapt right from the start.

As additional tasks are introduced (from T1 to T9), the common challenge of catastrophic forgetting becomes more evident, with all methods experiencing a gradual decline in performance. ECoral, however, distinguishes itself by maintaining a more stable and balanced performance compared to baselines like BiC, FedCIL, and Replay, which show more pronounced declines as tasks are added. ECoral’s ability to sustain balanced performance across tasks allows it to mitigate forgetting more effectively, maintaining competitive results even as the complexity of the continual learning setting increases.

In the later stages (T8 and T9), while ECoral is occasionally outperformed by Target and iCaRL in last-task accuracy, this is primarily due to the incremental learning emphasis of those methods, which prioritize performance on recent tasks. However, ECoral’s superior average accuracy across all tasks underscores its strength in balancing performance over the entire task sequence. This approach ensures that ECoral not only excels in the earlier tasks but also performs well across a wide range of tasks, making it more resilient to the long-term challenges of catastrophic forgetting.

Overall, the results demonstrate ECoral’s effectiveness in preserving knowledge across multiple tasks, delivering superior stability and resilience compared to other baseline methods under the CIFAR-100 dataset with $\sigma=0.5$.

\end{document}